\documentclass[]{spie}  %>>> use for US letter paper
\usepackage{comment}
\usepackage{booktabs}
\usepackage{amsmath,amsfonts,amssymb}
\usepackage{graphicx}
\usepackage{gensymb}
\usepackage{caption}
\usepackage{subcaption}
\usepackage{enumitem}
\usepackage[colorlinks=true, allcolors=blue]{hyperref}
\usepackage{listings}
\title{Class-specific diffusion models improve military object detection in a low-data domain}

\author[a]{Ella P. Fokkinga}
\author[a]{Jan Erik van Woerden}
\author[a]{Thijs A. Eker}
\author[a]{Sebastiaan P. Snel}
\author[a]{Elfi I.S. Hofmeijer}
\author[a]{Klamer Schutte}
\author[a]{Friso G. Heslinga}
\affil[a]{TNO - Intelligent Imaging, Oude Waalsdorperweg 63, the Hague, the Netherlands}
\authorinfo{Corresponding author: Friso G. Heslinga. E-mail: fgheslinga@gmail.com}

\begin{document} 
\maketitle

\begin{abstract}
Diffusion-based image synthesis has emerged as a promising source of synthetic training data for AI-based object detection and classification. In this work, we investigate whether images generated with diffusion can improve military vehicle detection under low-data conditions. We fine-tuned the text-to-image diffusion model FLUX.1 [dev] using LoRA with only 8 or 24 real images per class across 15 vehicle categories, resulting in class-specific diffusion models, which were used to generate new samples from automatically generated text prompts. The same real images were used to fine-tune the RF-DETR detector for a 15-class object detection task. Synthetic datasets generated by the diffusion models were then used to further improve detector performance. Importantly, no additional real data was required, as the generative models leveraged the same limited training samples. FLUX-generated images improved detection performance, particularly in the low-data regime (up to +8.0\% mAP$_{50}$ with 8 real samples). To address the limited geometric control of text prompt‑based diffusion, we additionally generated structurally guided synthetic data using ControlNet with Canny edge‑map conditioning, yielding a FLUX‑ControlNet (FLUX‑CN) dataset with explicit control over viewpoint and pose. Structural guidance further enhanced performance when data is scarce (+4.1\% mAP$_{50}$ with 8 real samples), but no additional benefit was observed when more real data is available. This study demonstrates that object-specific diffusion models are effective for improving military object detection in a low-data domain, and that structural guidance is most beneficial when real data is highly limited. These results highlight generative image data as an alternative to traditional simulation pipelines for the training of military AI systems.
\end{abstract}


% for each section one visual example and discussion why this adds value (for the limited amount of data)
% hier iets over hoe we literatuur hebben doorzocht/welke databases?

\section{Introduction}
\label{sec:intro} 

% BACKGROUND 
% 1. Why simulation? Standaard verhaal 
Deep learning–based object detection has become an important capability in military perception systems, supporting tasks such as target identification, threat assessment, and autonomous navigation \cite{Heslinga2024Simulation}. Developing robust detection models requires large, diverse, and well‑annotated datasets. However, in the military domain, access to operational imagery is often restricted, and data acquisition across the full range of relevant conditions, such as viewpoints, distances, environments, and configurations, is costly or infeasible. Synthetic data provides a complementary source to real imagery, enabling systematic control over scene composition and object appearance while reducing reliance on sensitive data.

% 2. GenAI-based => SOLUTION1
In recent years, generative AI (GenAI), particularly diffusion-based image synthesis, has emerged as a promising source of synthetic training data \cite{fokkinga2025generative}. Modern text-to-image diffusion models are capable of producing highly photorealistic and diverse images without requiring detailed 3D models or complex rendering setups. Prior work in non-military domains has shown that diffusion-generated imagery can improve downstream classification and detection performance \cite{azizi2023synthetic,chen2023diffusiondet}, suggesting that GenAI-based data generation may offer a flexible and scalable approach. % 3. Limited domain knowledge and need to finetune
Frontier diffusion models are trained on large amounts of image data, but have seen limited amounts of military scenes\cite{fokkinga2025generative}. Although these models are able to generate a generic photorealistic image of, for example, “a four-wheeled armed reconnaissance vehicle”, they lack knowledge about fine-grained vehicles classes and associated modular payloads. Fine-tuning techniques such as Low-Rank Adaptation (LoRA) \cite{hu2021lora} enable the injection of such particular domain knowledge to a diffusion model with a small set of examples. Typically, a small set of real-world examples is available for training an object detector, and this same set can also be used to fine-tune a generative model. Prior work has shown that fine-tuning a single diffusion model for a specific class yields realistic and flexible image generation \cite{soboleva2026tlora}. This suggests that object-specific diffusion models can produce high-quality synthetic data. However, it remains unclear to what extent such data improves downstream performance, particularly in specialized domains such as military object detection. 

% PROBLEM-2 => SOLUTION 2
Despite their flexibility and realism, diffusion models still face challenges in structural control. In text-to-image generation, attributes such as object orientation, viewpoint, and part configuration are typically specified only indirectly through natural-language prompts. Earlier simulation studies have shown that precise control over such geometric factors is critical for effective detector training \cite{Eker2023}. Yet, prompt-only diffusion models may not be able to provide the level of structural diversity and fidelity required for fine-grained military vehicle detection. To address this limitation, diffusion models can be augmented with explicit structural guidance \cite{zhang2023adding}, for example by conditioning the generative process on edge maps, providing direct control over outline and pose. Yet, the effectiveness of such explicit structural guidance in synthetic images to improve military object detectors has not been studied. 

% Approach
In this work, we explore to what extent GenAI-produced synthetic data can improve military object detection performance. Secondly, we investigate if additional explicit structural guidance of the synthetic data further enhances detector performance. To address these questions, we compare and mix three synthetic data sources: (i) object-specific images generated with the FLUX diffusion model, (ii) a structurally guided FLUX variant using ControlNet with edge‑map conditioning derived from 3D models, and (iii) physics-based 3D-model simulations, which we introduced in previous research \cite{Eker2023, Heslinga2024combining}. By evaluating these sources individually and in combination with limited real data, we assess both the value of GenAI‑generated synthetic data and the potential benefit of structural guidance for improving detector performance in low‑data military scenarios.

\section{Related works}
\label{sec:relatedworks}
\subsection{Synthetic data for military applications}
In the development of AI-algorithms for military applications, often only small and sensitive datasets are available. Many scenes show rare events, restricted assets, or controlled equipment, and public sources offer little variety. This limits the training of modern detection and segmentation models. Synthetic data offers a way to expand these datasets at low cost and with full control over content \cite{Heslinga2024Simulation}. Prior work shows clear gains in segmentation \cite{kulas2025unlockingthermalaerialimaging, truong20253dsmcos} and object detection  performance \cite{Eker2023, Eker2024fidelity, tang2024aerogen, FrankAlma2024}, when synthetic samples supplement real data. Several papers show that both the level of fidelity and variety of the simulations both influence the performance \cite{Eker2023, tang2024aerogen}. Military‑specific challenges such as occlusion and camouflage are particularly relevant in this context and have been investigated in recent work \cite{truong2025military}.

% 2. Physics-based for training AI => PROBLEM1
Physics-grounded, 3D model–based simulation pipelines have traditionally been the primary approach for generating synthetic training data for military vehicle detection. By rendering detailed 3D models in varied environments and explicitly controlling factors such as object pose, camera geometry, background, and illumination, these pipelines enable targeted coverage of relevant variation to train AI-models \cite{Eker2023}. 

\subsection{Diffusion models for generation of synthetic training data}
Diffusion models have gained broad interest in recent years as a powerful class of generative models. They operate by iteratively denoising a random noise sample through a sequence of learned steps to generate an image. Compared to generative adversarial networks (GANs), diffusion models offer more stable training dynamics and avoid issues such as mode collapse \cite{goodfellow2014generativeadversarialnetworks, chen2024implicitmultispectraltransformerlightweight, ozkanouglu2022infragan, Liu2021CycleGAN}. Recent text‑to‑image models, such as FLUX \cite{flux2024}, are capable of producing highly realistic images with strong structural consistency. These models follow text prompts with high fidelity and enable the generation of a wide variety of images from the same prompt by varying the initial noise input.

Several papers have demonstrated that diffusion-generated samples can improve downstream real-world performance on computer vision tasks \cite{tang2024aerogen, truong2025military, snel2025data}. However, these models are typically trained on large-scale, general-domain datasets and may not adequately capture domain-specific characteristics. When the target domain deviates from the training distribution, failures in generation start to increase, exhibiting inconsistencies in object geometry, scale, and viewpoint. This domain gap limits the direct applicability of off-the-shelf diffusion models in military contexts. Fine-tuning these models on a small set of domain-specific examples can help mitigate these issues, improving both structural consistency and visual realism.

\subsection{Adapting diffusion models for military applications}
Fine-tuning diffusion models on small, domain-specific datasets can substantially improve generation quality. However, due to the large number of parameters in modern diffusion models, full fine-tuning is often infeasible. To address this, lightweight adaptation methods have been introduced that approximate the performance of full fine-tuning while requiring significantly fewer trainable parameters \cite{fokkinga2025generative}. One such method is LoRA \cite{hu2021lora}, which injects trainable low-rank updates into selected layers of the model. This enables efficient fine-tuning using limited data while keeping the majority of the model weights fixed.

Additionally, recent work introduces modules that guide diffusion models with clear structure. ControlNet \cite{zhang2023adding} is a well-known example, which conditions image generation on inputs such as injects edge maps, depth maps, or segmentation layouts. By injecting this structural information into the denoising process, ControlNet enforces the global shape while allowing the base model to fill texture and appearance.

Research in diffusion‑based synthetic data for military applications remains limited. Most public work focuses on consumer photography or artistic images. Few studies analyse downstream performance in detection tasks that match military needs. This gap motivates our study, using a modern text‑to‑image models like FLUX \cite{flux2024} combined with structural control modules such as ControlNet \cite{zhang2023adding}.

\section{Methods}
\label{sec:methods}
In this study, four data sources were used to train the downstream object detector: (i) real-world images, (ii) synthetic images generated using object-specific diffusion models (\textit{FLUX}), (iii) structurally guided diffusion images using ControlNet (\textit{FLUX-CN}), and (iv) 3D model–based simulation data. The 3D-model simulations served a dual role: they were used both as an independent synthetic dataset for detector fine-tuning and as a source of structural guidance (via Canny edge maps) in the \textit{FLUX-CN} pipeline. In the following sections, each dataset is described.

Two stages of model adaptation can be distinguished: (1) fine-tuning the diffusion model using LoRA for synthetic data generation, and (2) fine-tuning the object detector (RF-DETR) using real and synthetic datasets. The methods of stage (1) are described in Section \ref{methods:datasets:flux-base} and \ref{methods:datasets:flux-cn}, and those for stage (2) in Section \ref{methods:object_detection}.

\subsection{Real dataset}
\label{methods:datasets:real}
The real‑world dataset used in this study is identical to the dataset employed in earlier work on fine‑grained military vehicle detection \cite{Heslinga2024combining}. In total, 959 real RGB images were collected via web scraping for 15 military vehicle classes. An overview of the classes is provided in Table \ref{tab:vehicles}. The dataset was previously divided into 360 training images (corresponding to 24 images per class), 150 validation images (10 per class), and 449 test images (21–50 samples per class). We constructed two training subsets: 8 and 24 real images per class, which we used both as standalone fine-tuning data for the object detector and as conditioning samples for diffusion‑model fine‑tuning (Section \ref{methods:datasets:flux-base}). 

\begin{table}[tb!]
\caption{Overview of the military vehicle classes.} 
\label{tab:vehicles}
\centering
\begin{tabular}{@{}ll@{}}
\toprule
Superclass                 & Classes                               \\ \midrule
Armoured personnel carrier & Boxer, BTR-80, TPz Fuchs, Patria      \\
Scout car                  & Fennek, BRDM-2                        \\
Battle tank                & Leopard, M1 Abrams, T90, CV90         \\
Howitzer                   & M109, 2S19 Msta, Panzerhaubitze 2000  \\
Military truck             & DAF YA 4440, Scania                   \\ \bottomrule
\end{tabular}
\end{table}

\subsection{Synthetic dataset generation}
\subsubsection{Object-specific diffusion (FLUX)}
\label{methods:datasets:flux-base}
\begin{figure}[tb!]
    \centering
    \includegraphics[width=0.9\linewidth]{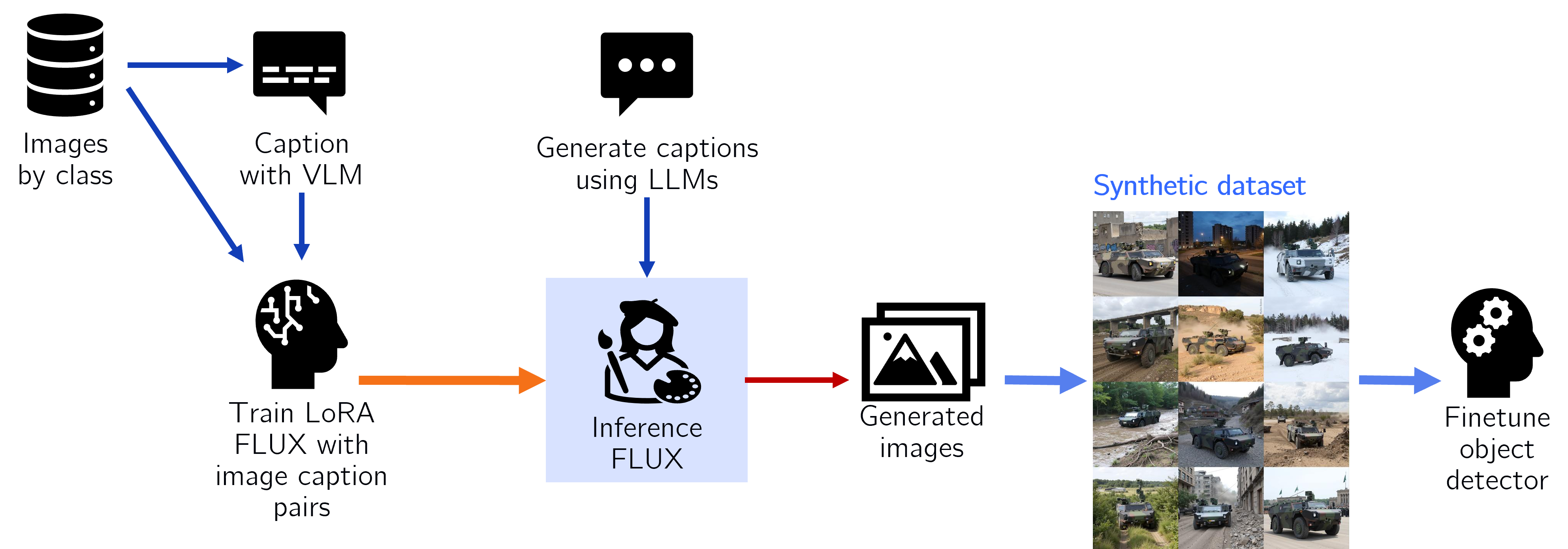}
    \caption{The base workflow for image-synthesis using diffusion models and image-caption pairs. VLM = Vision Language Model, LLM = Large Language Model, FLUX = a text-to-image diffusion model.}
    \label{fig:pipeline-base}
\end{figure}

Class‑specific synthetic training data was generated using a diffusion pipeline based on the text-to-image model FLUX.1 [dev] \cite{flux2024}. Before synthetic images could be generated, the base diffusion model first had to be adapted to the military domain. Off‑the‑shelf diffusion models lack exposure to specialized military vehicle concepts and therefore do not reliably produce class‑consistent or structurally accurate vehicles \cite{fokkinga2025generative}. To address this, we fine‑tuned FLUX using LoRA \cite{hu2021lora}, producing separate object‑specific models for each of the 15 classes using the limited set of available real images (8 or 24 per class). The process consisted of (i) captioning real images, (ii) LoRA‑based fine‑tuning of FLUX per class, (iii) prompt generation, (iv) synthetic image generation, (v) automatic annotation, and (vi) downstream use in object detector fine-tuning. This workflow is illustrated in Figure~\ref{fig:pipeline-base}.

\begin{figure}[tb!]
    \centering
    \vspace{0.8cm}
    \includegraphics[width=\linewidth]{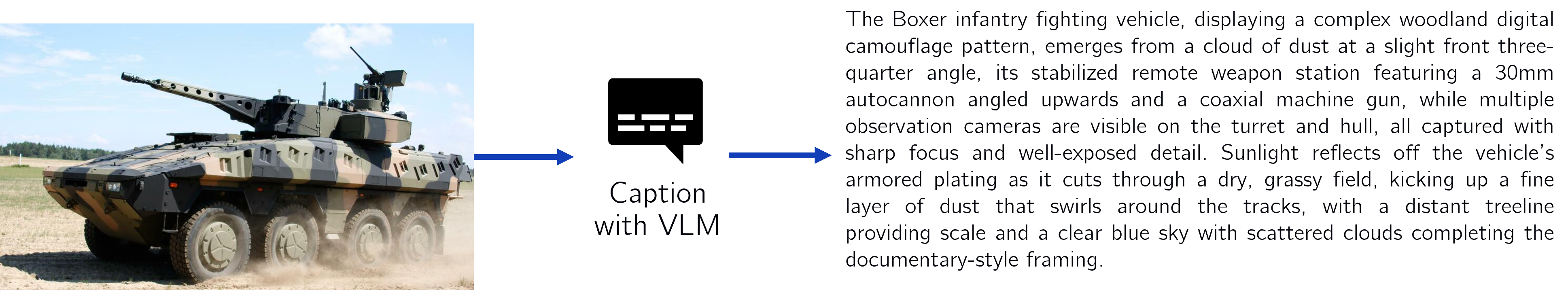}
    \caption{Example of a real image of a Boxer and the caption to describe it. Such image-caption pairs are inputs for fine-tuning a  FLUX diffusion model.}
    \label{fig:example-captioning}
\end{figure}

For each real image (as described in Section \ref{methods:datasets:real}), a structured caption was generated using the vision‑language model (VLM) Gemma‑3‑12b‑it \cite{team2025kimi}. The system and user prompt used to generate these captions are reported in the Appendix (Section \ref{appendix:prompts}). Captions described vehicle appearance, environmental conditions, and contextual attributes. The VLM was instructed to write captions containing details about vehicle appearance, environment, weather and operational context. An example of a caption generated by the VLM and the corresponding image is provided in Figure \ref{fig:example-captioning}.

FLUX was fine-tuned using LoRAs attached to the model’s attention layers \cite{hu2021lora}. A separate LoRA was trained for each class using its corresponding image-caption pairs, while the base FLUX weights remain frozen. LoRA parameters were set to a linear rank of 32 and linear alpha of 32, with full-rank adaptation enabled. Fine-tuning was conducted for 2,000 steps with a batch size of 1 and gradient accumulation set to 1. Both the U-Net and text encoder components were trained with gradient checkpointing enabled to reduce memory footprint. The optimizer was AdamW (8-bit), with a learning rate of 0.0004 and weight decay of 0.0001. The hyperparameters are either the default settings or were selected based on preliminary experiments to ensure stable training and consistent generation quality. To analyse the impact of the number of real images used to fine-tune FLUX, two set sizes were considered: 8 and 24 real images. To diversify scene composition and phrasing beyond the initial captions generated by the VLM, we generated new captions using GPT-4. The system and user prompt to generate these captions are reported in the Appendix (Section \ref{appendix:prompts}). With the class‑specific LoRA applied, 150 images per class are synthesized.

All generated images were annotated automatically using GroundingDINO\cite{liu2024grounding} in zero‑shot, open‑vocabulary mode. GroundingDINO was prompted with “military vehicle”, and the top‑1 bounding box was retained. The class label was not predicted by GroundingDINO, but instead assigned based on the class-specific LoRA used during image generation. This produced a single bounding box per image and a class label consistent with synthesis conditions. Visual inspection indicated that the generated annotations were well-aligned with the synthesized objects.

\subsubsection{Diffusion with structural guidance (FLUX-CN)}
\label{methods:datasets:flux-cn}
A second set of synthetic images was generated using FLUX combined with ControlNet, which introduces explicit structural conditioning into the diffusion process \cite{zhang2023adding, fokkinga2025generative}. To construct these structural guidance maps, Canny edges were extracted from 3D vehicle models (Section \ref{methods:datasets:3dmodel}), which were also used as a separate synthetic dataset in our experiments. The 3D models were rendered in Blender \cite{Blender2018} across a variety of viewpoints and distances, matching the axes of simulation variation known to be important from earlier work \cite{Eker2023, Heslinga2024combining}. These included viewpoint changes, object rotation, and camera distance. The edge maps encoded the vehicle's shape, spatial layout, and viewpoint.

The same set of captions used for the FLUX-generated dataset, generated by GPT-4 (see Appendix \ref{appendix:prompts} for prompts), was reused to generate new synthetic images. However, explicit geometric descriptions (e.g., viewpoint and orientation) were removed from the prompts, as this information was now provided through the structural guidance of the edge maps. During synthesis, the same LoRAs that were trained for the base FLUX pipeline were reused. ControlNet was applied solely during the generation stage. Bounding box annotations were reused from the 3D model–based simulations from which the edge maps were derived. Because these edge maps guided image generation, the resulting synthetic images remained geometrically aligned with the original simulated objects. For each class and fine‑tuning regime (8 or 24 real examples), 150 synthetic images were generated.

\begin{figure}[tb!]
    \centering
    \begin{subfigure}[t]{0.32\textwidth}
        \centering
        \includegraphics[width=\linewidth]{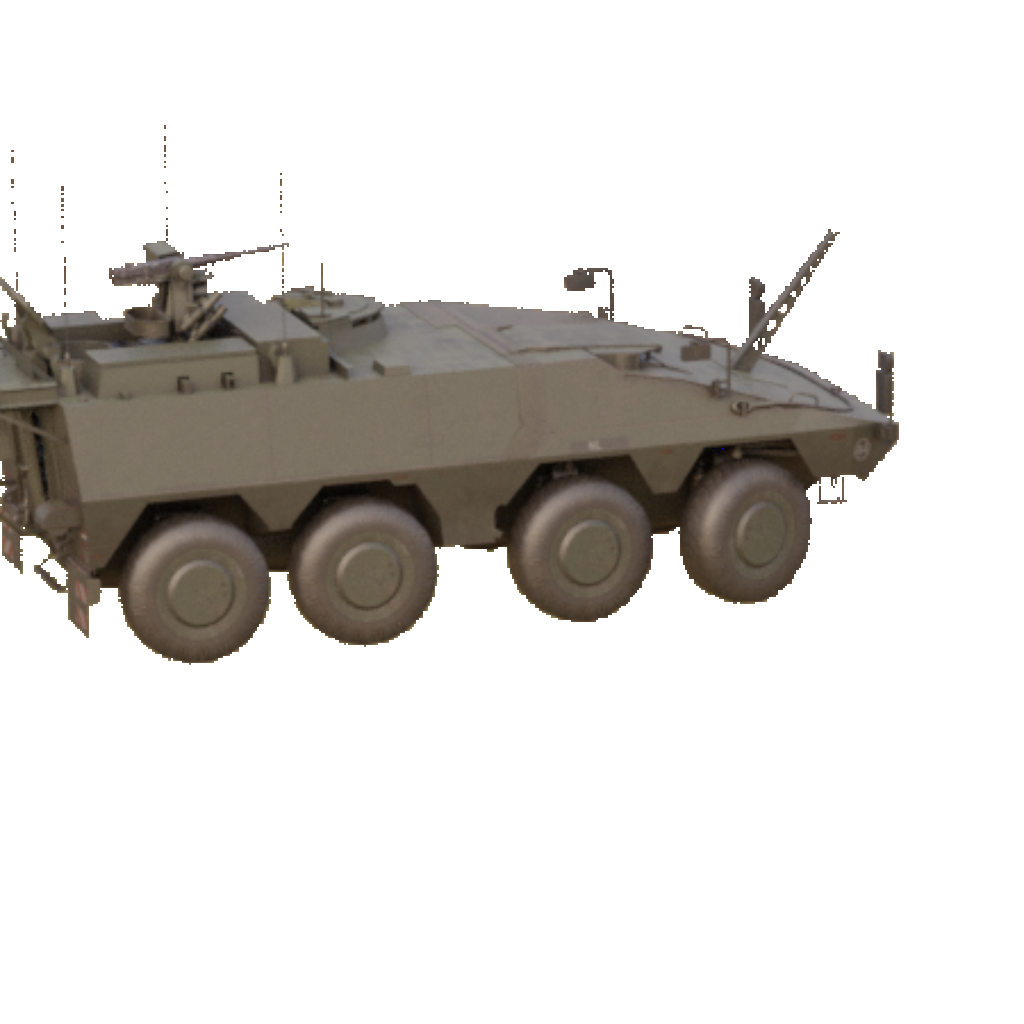}
        \caption{3D model in Blender}
        \label{fig:a_image}
    \end{subfigure}
    \hfill
    \begin{subfigure}[t]{0.32\textwidth}
        \centering
        \includegraphics[width=\linewidth]{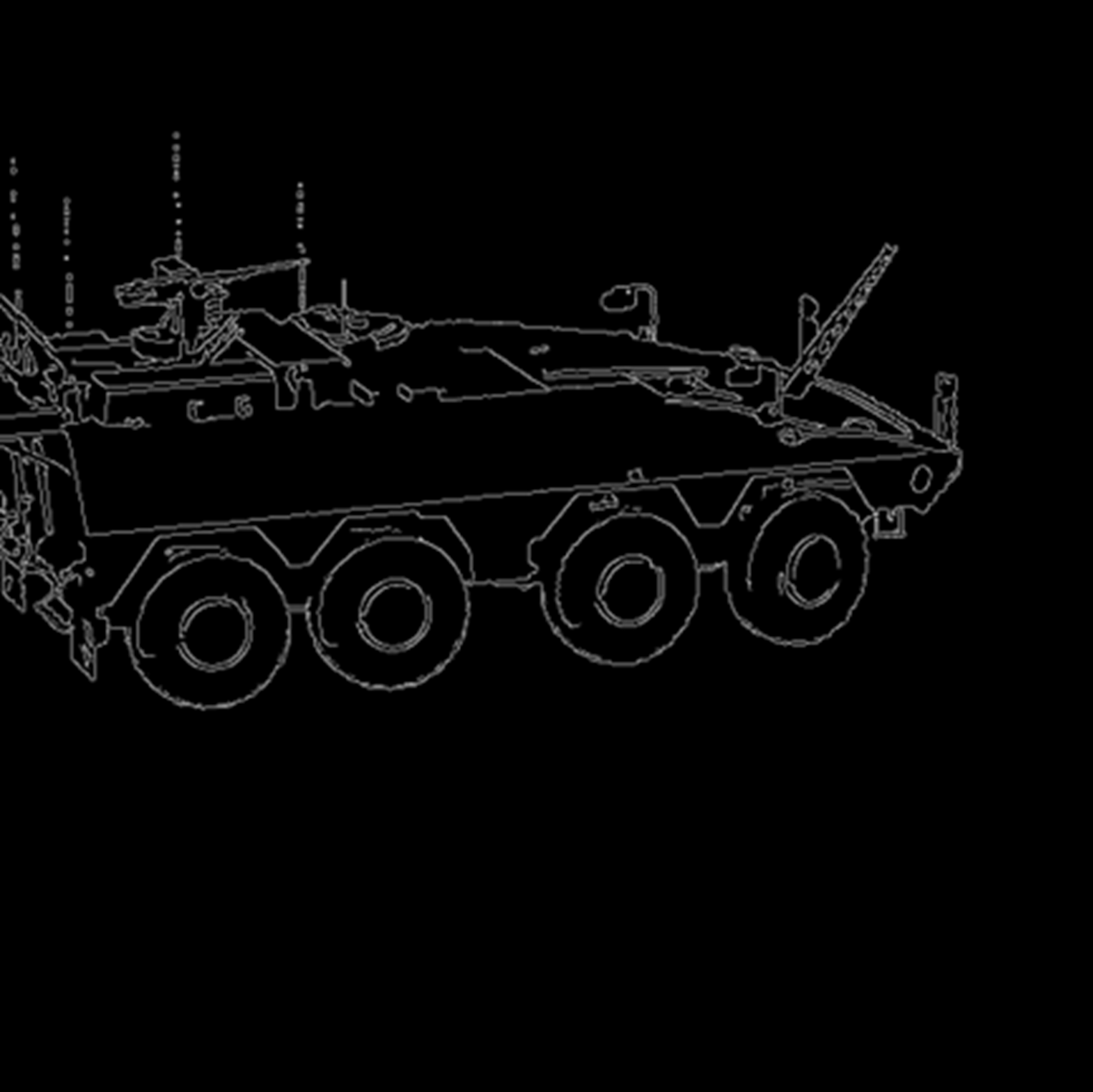}
        \caption{Canny edge map}
        \label{fig:b_canny_fg}
    \end{subfigure}
    \hfill
    \begin{subfigure}[t]{0.32\textwidth}
        \centering
        \includegraphics[width=\linewidth]{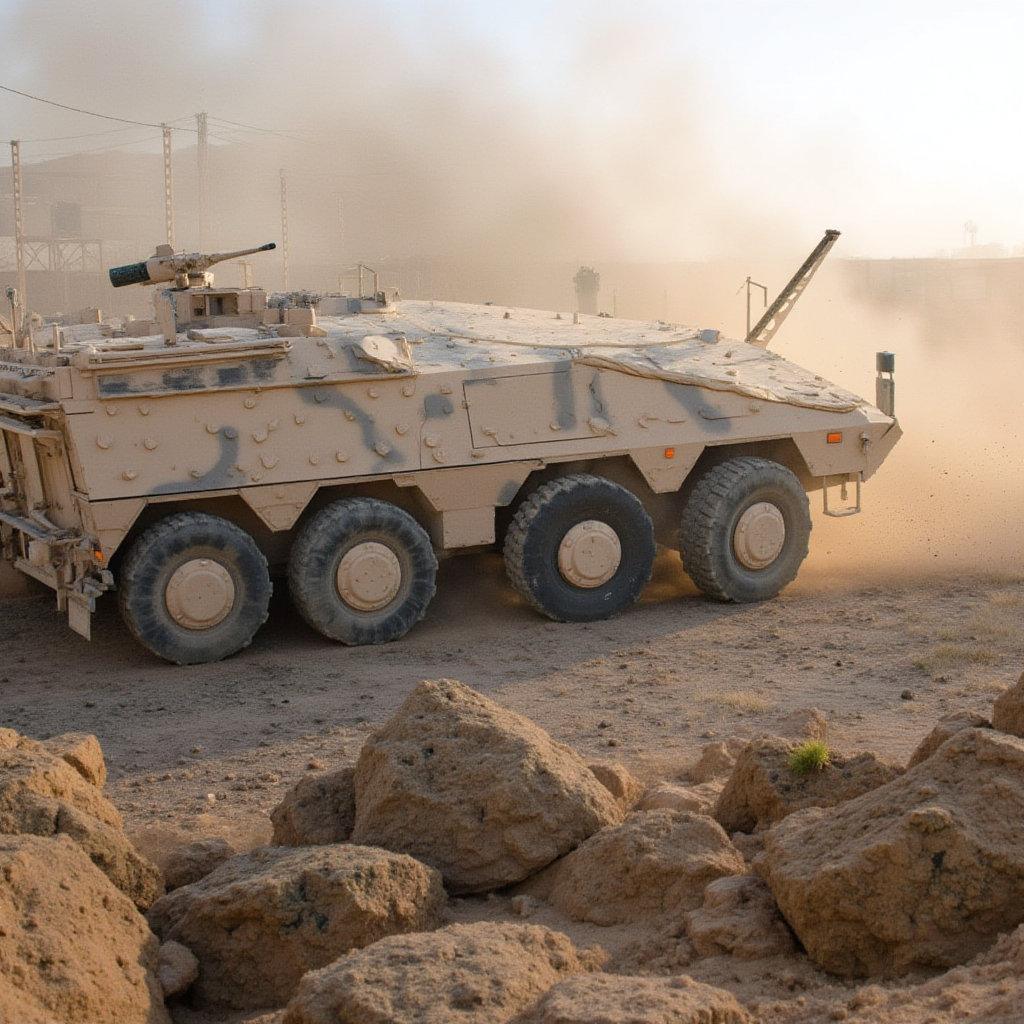}
        \caption{Generated with FLUX-ControlNet}
        \label{fig:c_flux_cn}
    \end{subfigure}

    \caption{Illustration of the structural‑guidance process used in FLUX‑ControlNet image synthesis for a Boxer. (a) A Blender‑rendered simulation image provides a precise geometric reference for vehicle pose, viewpoint, and component configuration. (b) A Canny edge map is extracted from the simulated image in (a), capturing structural information that textual prompts do not reliably enforce. (c) A FLUX‑ControlNet generated image conditioned on both the class‑specific LoRA and the edge map. The prompt used for this example was: \textit{“A Boxer IFV covered in dust-caked, patchy camouflage, sensors and turret barely visible above a cluster of boulders. Swirling dust in the air from a passing support truck, sunlight filtering through the haze.”}}
    \label{fig:workflow_guidance_controlnet}
\end{figure}

\subsubsection{3D model–based simulations}
\label{methods:datasets:3dmodel}
In addition to diffusion-based data generation, we included the 3D model–based simulation dataset as a complementary source of synthetic training data. This dataset was derived from the same simulation pipeline described by Eker et al. \cite{Eker2023}. The simulation pipeline was based on Blender‑rendered \cite{Blender2018} 3D vehicle models placed in HDRI scenes, sampled across a wide range of simulation parameters including viewpoint variation, model textures, object–camera distance, yaw, pitch, roll, and model configurations. For more details, we refer to Eker et al. (2023) \cite{Eker2023}. For this study, a subset of 2,250 synthetic images (150 per class) was randomly selected from this simulation pipeline. The same set of 150 synthetic images per class was used to extract Canny edge maps that served as structural guidance for the FLUX‑ControlNet experiments described in the previous Section \ref{methods:datasets:flux-cn}.

\subsection{Object detection}
\label{methods:object_detection}
\subsubsection{Experimental setup}
The goal of our experiments was to quantify the contribution of diffusion‑based synthetic data to object detection performance in a low‑data regime involving 15 military vehicle classes, and to assess if explicit structural guidance during image generation further enhanced the utility of such synthetic data. Detector performance was compared between models fine-tuned on real data and models augmented with different sources of synthetic data. Specifically, we evaluated synthetic data generated with \textit{FLUX}, structurally guided diffusion data generated with FLUX‑ControlNet (\textit{FLUX-CN}), and traditional 3D model–based simulation data (\textit{3d-model-sim}), both in isolation and in combination with real images. For all FLUX-based combinations, the number of real images (8 or 24 per class) matched the set used to fine-tune the corresponding diffusion models. 

\subsubsection{RF-DETR}
As object detection model, we employ the RF-DETR (Medium) framework \cite{rf-detr}, a fully transformer-based detector that builds upon the DETR paradigm. The model was initialized from pretrained weights, combining a DINOv2\cite{oquab2024dinov2learningrobustvisual} backbone trained on large-scale image data with a detection head pretrained on COCO. By leveraging the DINOv2 backbone for feature extraction, strong visual representations are provided that contribute to robust performance. The model has demonstrated state-of-the-art results in both in-domain and out-of-domain generalization compared to conventional detectors such as YOLOv11 \cite{jocher2024ultralytics}. Training was performed at an input resolution of $960 \times 960$. The model was trained for 80 epochs, using a batch size of 12, a learning rate of $1.0 \times 10^{-5}$ for the detection head, and a learning rate of $3.0 \times 10^{-5}$ for the backbone. These settings were selected to balance computational efficiency and detection performance. All remaining training hyperparameters followed the default settings of the RF‑DETR implementation. The real-image validation set described in Section~\ref{methods:datasets:real} was used to monitor training and select the best-performing model checkpoint. For experiments combining multiple data sources, a balanced sampling was approximated by adjusting their relative frequencies in the training set, resulting in equal contributions for each dataset.

\subsubsection{Evaluation}
Detector performance was quantified using mean average precision (mAP) \cite{Padilla2020}. Both mAP$_{50}$ and mAP$_{50{:}95}$ are reported, to capture performance under a lenient intersection over union (IoU) threshold (emphasizing classification) and under increasingly strict IoU thresholds (reflecting localisation accuracy), respectively. All models were evaluated on the same real‑image test set of 449 samples (21–50 images per class) as used in prior work \cite{Heslinga2024combining} (Section \ref{methods:datasets:real}). Each experiment was repeated three times with different random seeds, and we report the mean and standard deviation of mAP-metrics over the runs.

\section{Results}
\subsection{Evaluation of diffusion-based synthetic data}
\begin{figure}[tb!]
    \centering
     \begin{subfigure}[t]{0.24\textwidth}
        \centering
        \includegraphics[width=\linewidth]{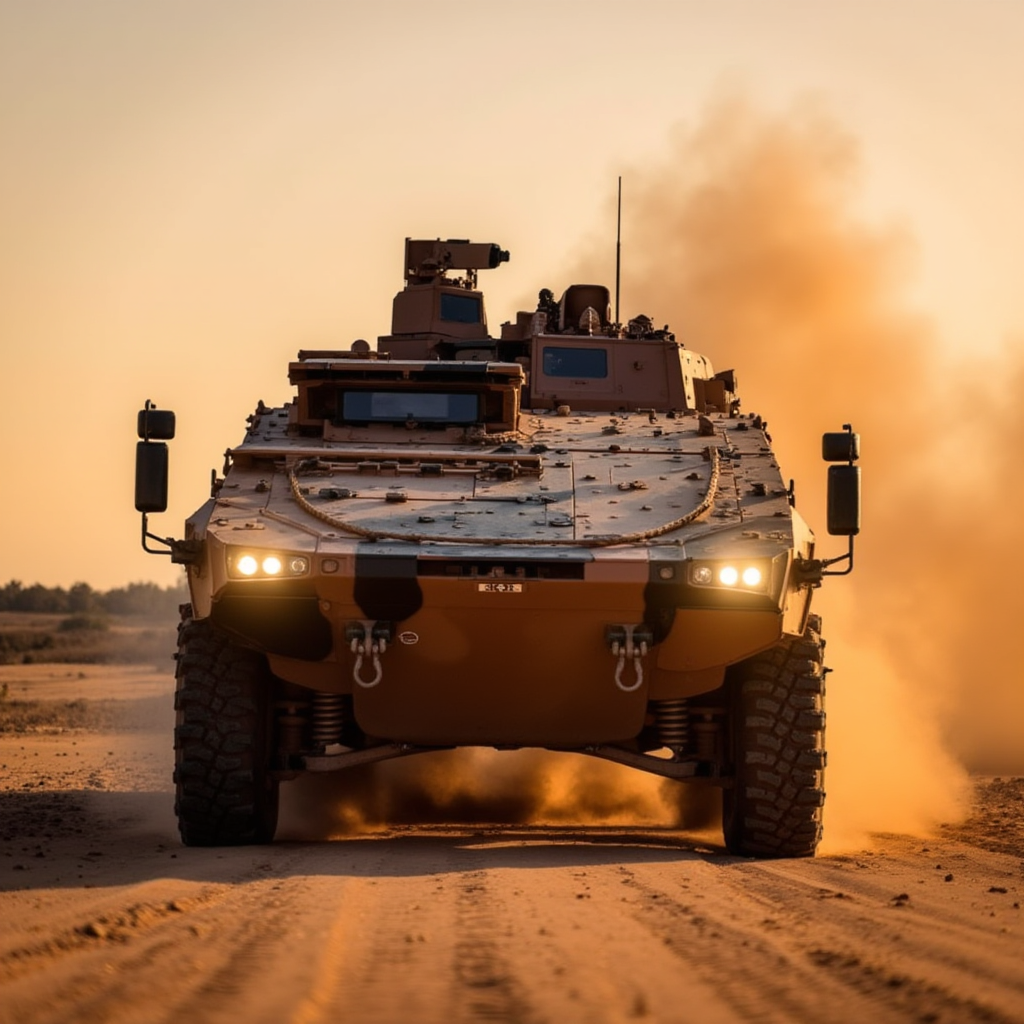}
        \caption{Boxer}
        \label{fig:boxer_flux}
    \end{subfigure}
    \hfill
    \begin{subfigure}[t]{0.24\textwidth}
        \centering
        \includegraphics[width=\linewidth]{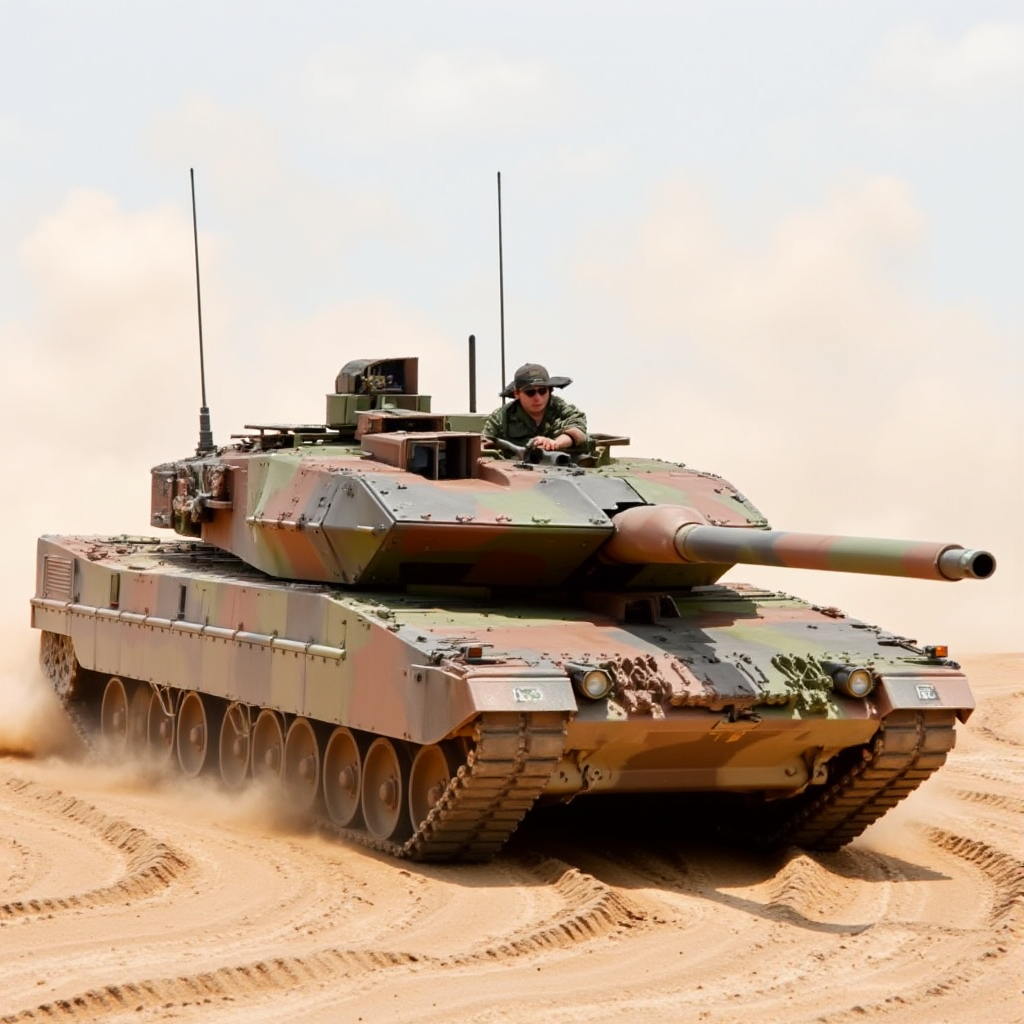}
        \caption{Leopard}
        \label{fig:leopard_flux}
    \end{subfigure}
    \hfill
    \begin{subfigure}[t]{0.24\textwidth}
        \centering
        \includegraphics[width=\linewidth]{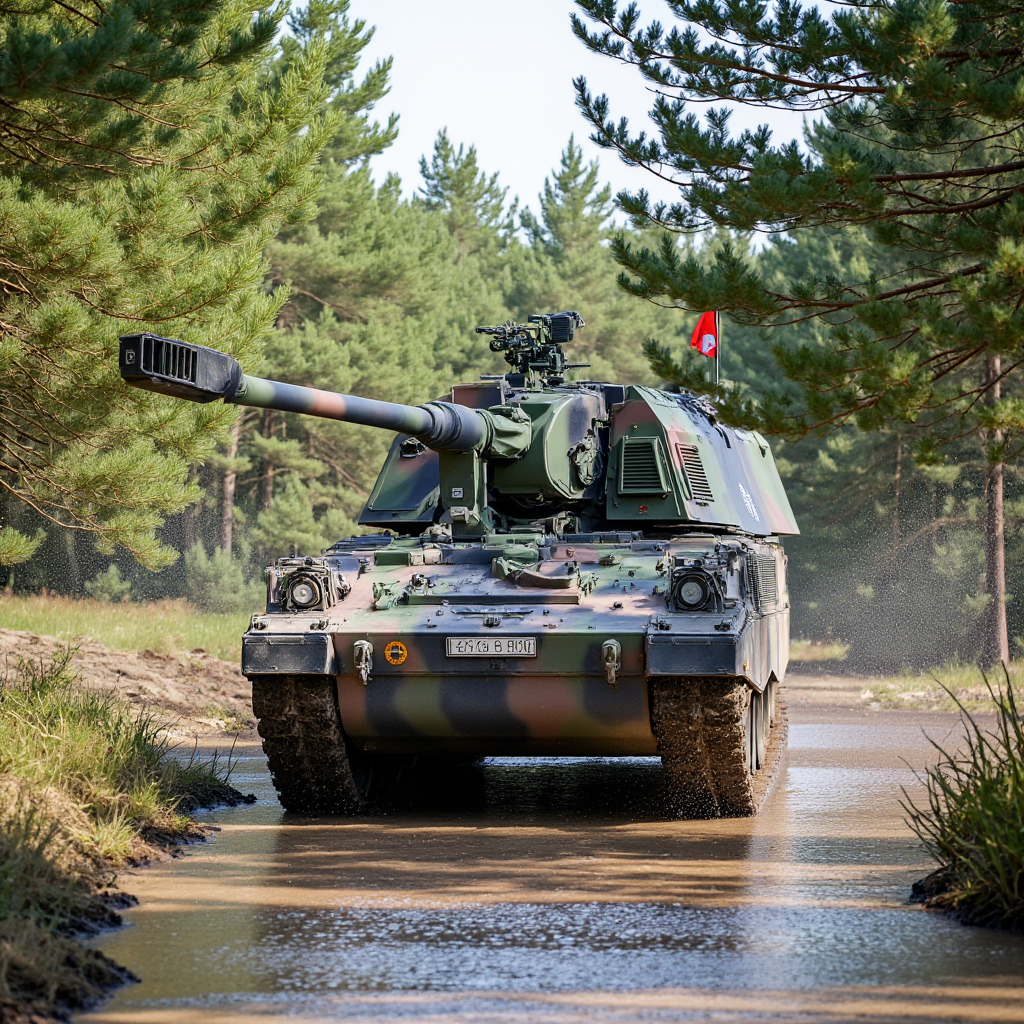}
        \caption{Panzerhaubitze 2000}
        \label{fig:howitzer}
    \end{subfigure}
    \hfill
     \begin{subfigure}[t]{0.24\textwidth}
        \centering
        \includegraphics[width=\linewidth]{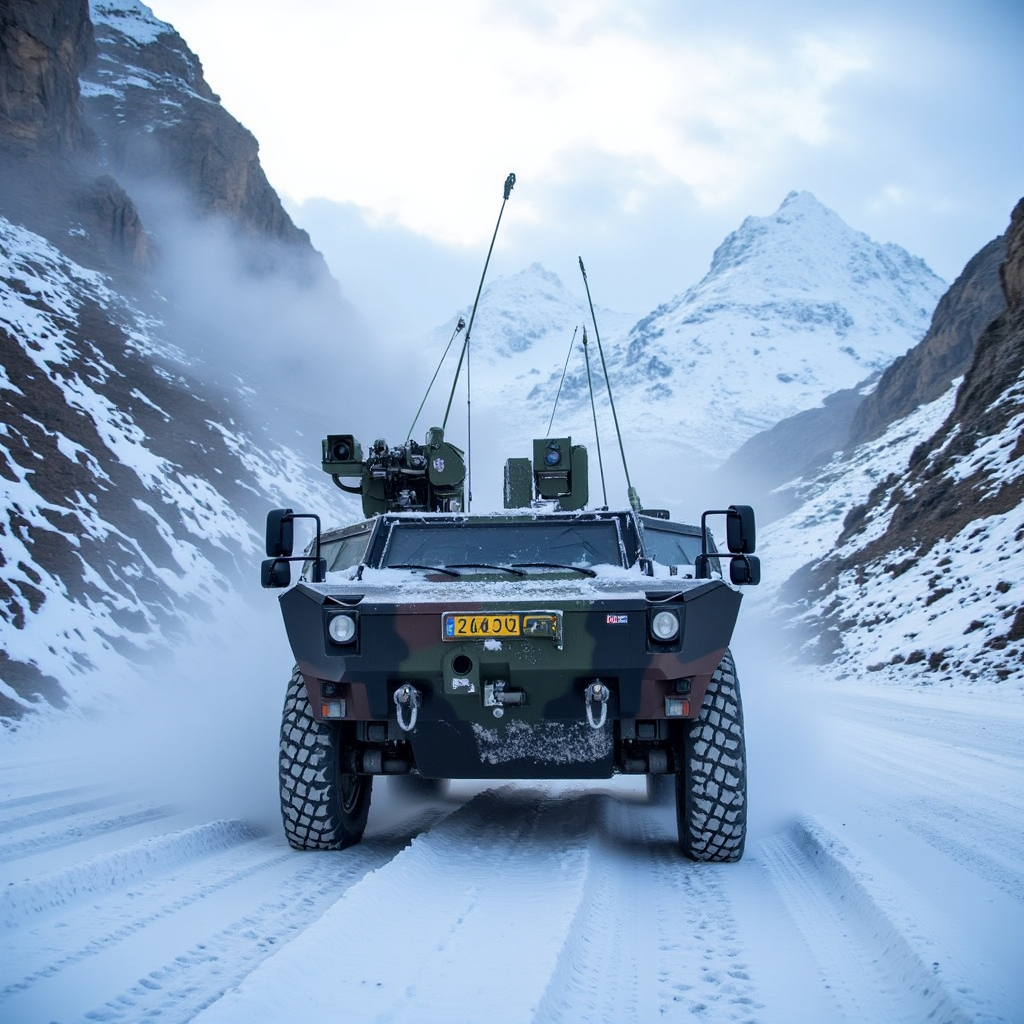}
        \caption{Fennek}
        \label{fig:fennek}
    \end{subfigure}
    \caption{Examples of images generated by the object-specific FLUX-models for four class, each fine-tuned with 24 real samples per class.}
    \label{fig:examples_flux}
\end{figure}

Figure \ref{fig:examples_flux} shows four representative samples generated by the object‑specific FLUX models, each fine‑tuned using 24 real images of the corresponding class. The examples highlight that FLUX is able to synthesize realistic vehicle appearances with substantial environmental diversity.

\begin{figure}[tb!]
    \centering
    \includegraphics[width=\linewidth]{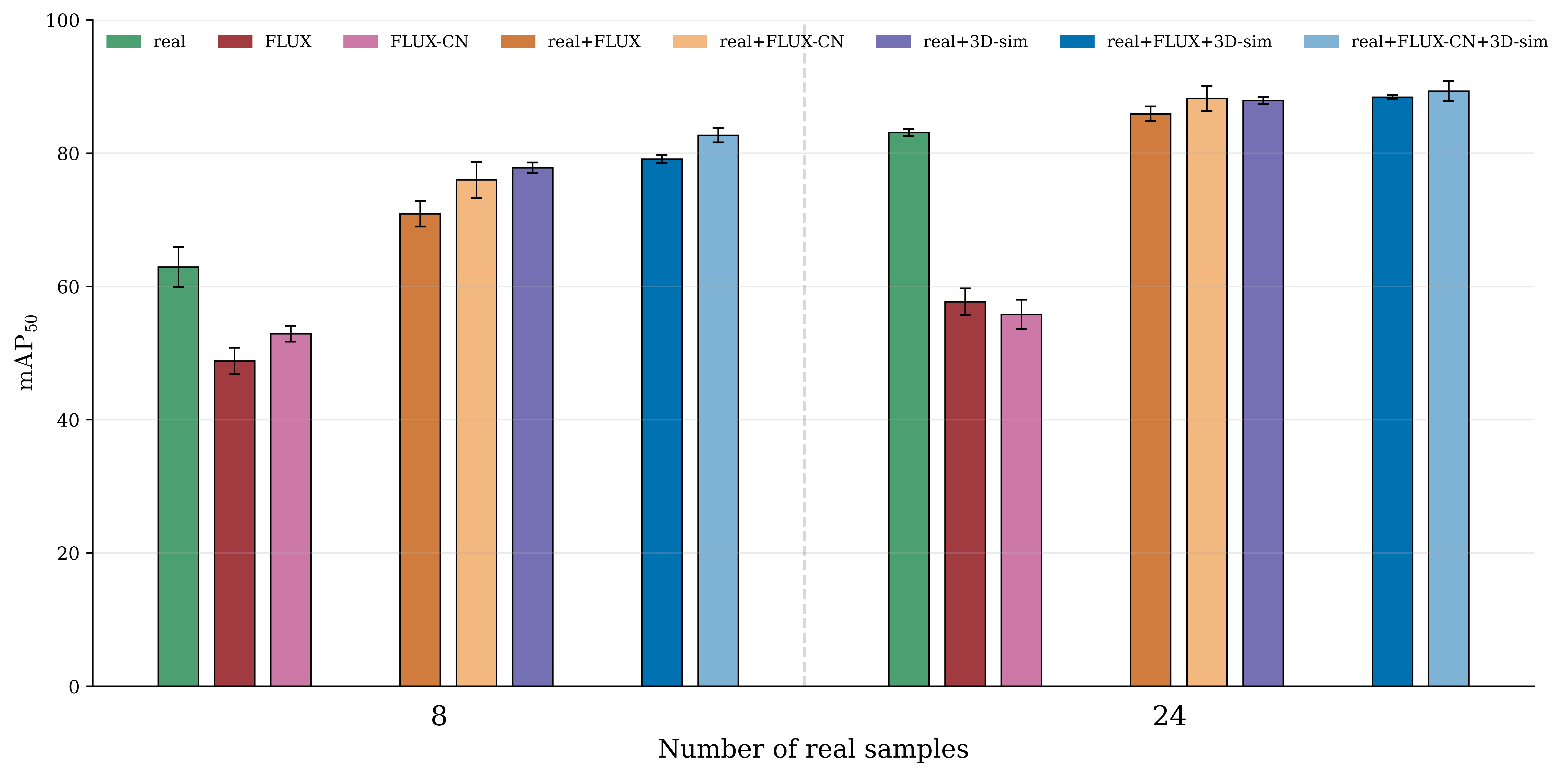}
    \caption{Comparison of object detection performance (mAP$_{50}$) on the test set for RF-DETR trained with different combinations of real and synthetic data. Results are shown for 2 data regimes: 8 and 24 real samples. Synthetic data generated with FLUX is obtained using a FLUX model fine-tuned with LoRA on the corresponding number of real samples. FLUX-CN denotes the structurally guided variant, where image generation is conditioned using ControlNet with Canny edge maps derived from 3D models. Error bars indicate standard deviation over three runs.}
    \label{fig:mAP_barchart}
\end{figure}

% Experiments:
% Does GenAI add value? 
Table \ref{tab:mAP_flux} reports the object detection performance of RF-DETR on the real test data, when fine-tuned on the three types of datasets: real, \textit{FLUX}, and \textit{3d-model-sim}. Comparing the two metrics, $\text{mAP}_{50}$ and $\text{mAP}_{50:95}$, we observe that the performance gap between the two metrics remains consistent across all settings. Figure \ref{fig:mAP_barchart} provides a visual summary of the $\text{mAP}_{50}$ results across the different training configurations. Fine-tuning the detector solely on 3D model–based synthetic data results in a low performance ($\text{mAP}_{50}=26.6$), consistent with earlier findings \cite{Heslinga2024combining}. In contrast, a small amount of real data yields a substantial increase in performance: using 8 real samples results in $\text{mAP}_{50}=62.9$, and expanding this to 24 samples raises performance further to $\text{mAP}_{50}=83.1$. 

\begin{table}[tb!]
\centering
\caption{Object detection performance on the test set for RF-DETR fine-tuned on the different (combinations of) datasets. FLUX refers to synthetic data generated using a FLUX model fine-tuned with LoRA on the corresponding number of real samples.}
\begin{tabular}{lllll}
\hline
                                & \multicolumn{2}{c}{\textbf{$\text{mAP}_{50}$ [$\pm$ std]}}                   & \multicolumn{2}{c}{\textbf{$\text{mAP}_{50:95}$ [$\pm$ std]}}                \\ \hline
3d-model-sim               & \multicolumn{2}{c}{26.6 [4.0]} & \multicolumn{2}{c}{24.2 [4.0]} \\ \hline
\textit{number of real samples} & \multicolumn{1}{c}{\textit{8}} & \multicolumn{1}{c}{\textit{24}} & \multicolumn{1}{c}{\textit{8}} & \multicolumn{1}{c}{\textit{24}} \\ \hline
real only                  & 62.9 [3.0]     & 83.1 [0.5]    & 57.4 [2.6]     & 77.1 [0.8]    \\
FLUX                       & 48.8 [2.0]     & 57.7 [2.0]    & 42.8 [1.7]     & 51.1 [1.6]    \\
real + 3d-model-sim        & 77.8 [0.8]     & 87.9 [0.5]    & 71.3 [0.9]     & 81.8 [0.4]    \\
real + FLUX                & 70.9 [1.9]     & 85.9 [1.1]    & 64.6 [1.8]     & 79.9 [1.1]    \\
real + FLUX + 3d-model-sim & 79.1 [0.6]     & 88.4 [0.3]    & 72.9 [0.8]     & 82.5 [0.1]    \\ \hline
\end{tabular}
\label{tab:mAP_flux}
\end{table}

For both the 8- and 24-sample settings, fine-tuning RF-DETR on \textit{FLUX} data underperforms fine-tuning on the corresponding real samples. The performance gap increases with more real data, rising from approximately 22\% to 31\% in $\text{mAP}_{50}$. However, combining real images with \textit{FLUX} data improves detection performance. For 8 real samples, adding \textit{FLUX} data to the fine-tuning set of RF-DETR increases performance by approximately 8\% in $\text{mAP}_{50}$. When 24 real samples are available, the gain is smaller, around 3\%.
 
% The performance gain from adding \textit{FLUX} data is smaller than that obtained from incorporating \textit{3d-model-sim}, with the latter consistently yielding higher improvements. Combining all three sources provides a modest additional gain over the best two-source combination. Specifically, an improvement of approximately 2\% in $\text{mAP}_{50}$ is observed for the 8-sample setting, and less than 1\% for the 24-

\subsubsection{Evaluation of structural guidance}
The prompt used to generate the Boxer in Figure \ref{fig:boxer_flux} was: \textit{“A Boxer IFV, covered in a dust-caked, patchy camouflage, is presented in a partially obscured, front three-quarter view behind a cluster of boulders. The vehicle’s sensors and turret barely peek above the rocks as swirling dust is kicked up by a passing support truck, with sunlight filtering through the haze and lending a sense of urgency to the scene.”}. As illustrated in Figure \ref{fig:boxer_flux}, the generated image does not adhere to the geometric constraints described in the prompt. In particular, the specified viewpoint (e.g., \textit{“front three-quarter view”}) and partial occlusion are not consistently reflected in the output. This behavior is observed across many generated samples, where the model tends to default to perspectives such as frontal or slightly side-angled views, despite more detailed positional instructions in the prompt. 

\begin{figure}[tb!]
    \centering
     \begin{subfigure}[t]{0.3\textwidth}
        \centering
        \includegraphics[width=\linewidth]{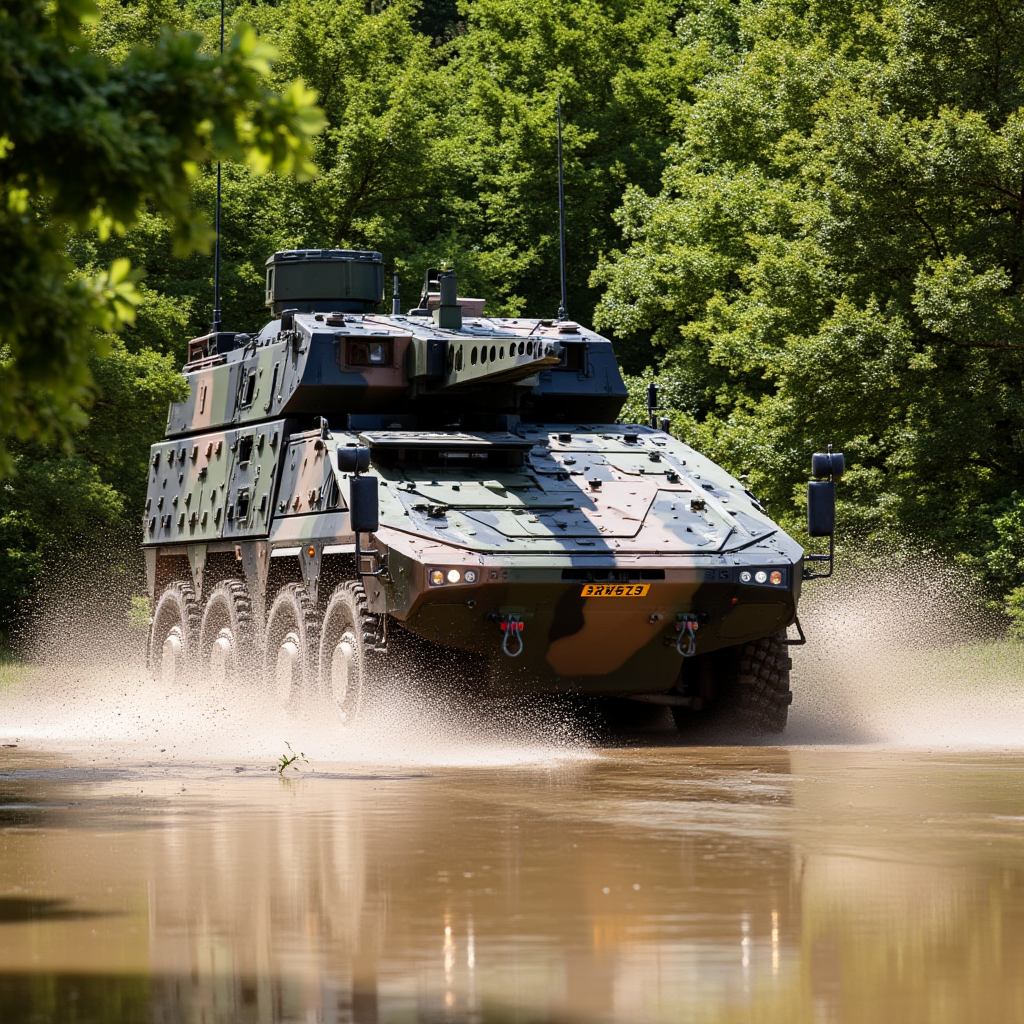}
        \caption{Generated by FLUX}
        \label{fig:boxer2_flux}
    \end{subfigure}
    \hfill
    \hfill
    \begin{subfigure}[t]{0.3\textwidth}
        \centering
        \includegraphics[width=\linewidth]{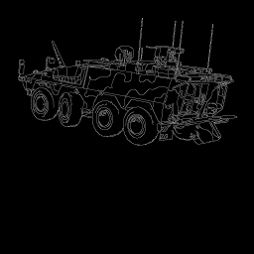}
        \caption{3D model–based edge map}
        \label{fig:boxer2_cannyfg}
    \end{subfigure}
    \begin{subfigure}[t]{0.3\textwidth}
        \centering
        \includegraphics[width=\linewidth]{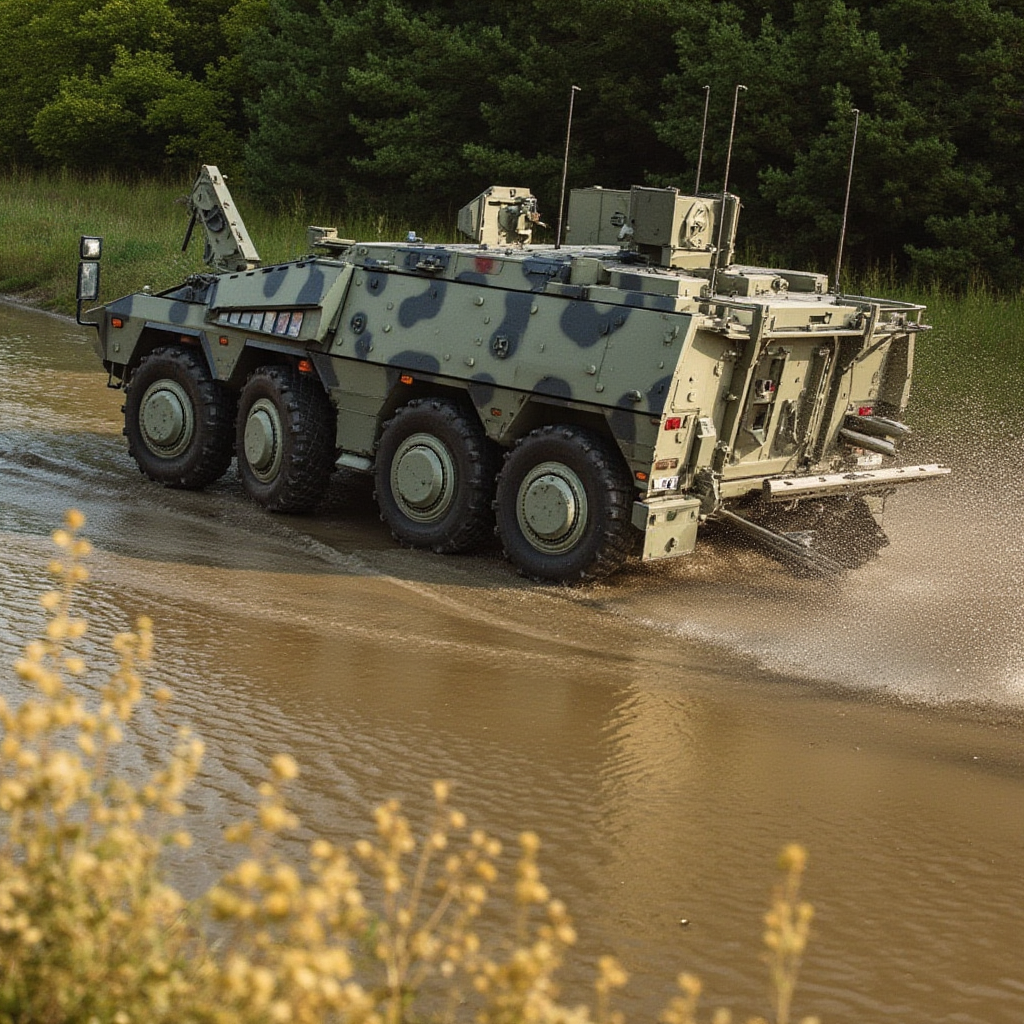}
        \caption{Generated by FLUX-CN}
        \label{fig:boxer2_flux_cn}
    \end{subfigure}
    \hfill
    \begin{subfigure}[t]{0.3\textwidth}
        \centering
        \includegraphics[width=\linewidth]{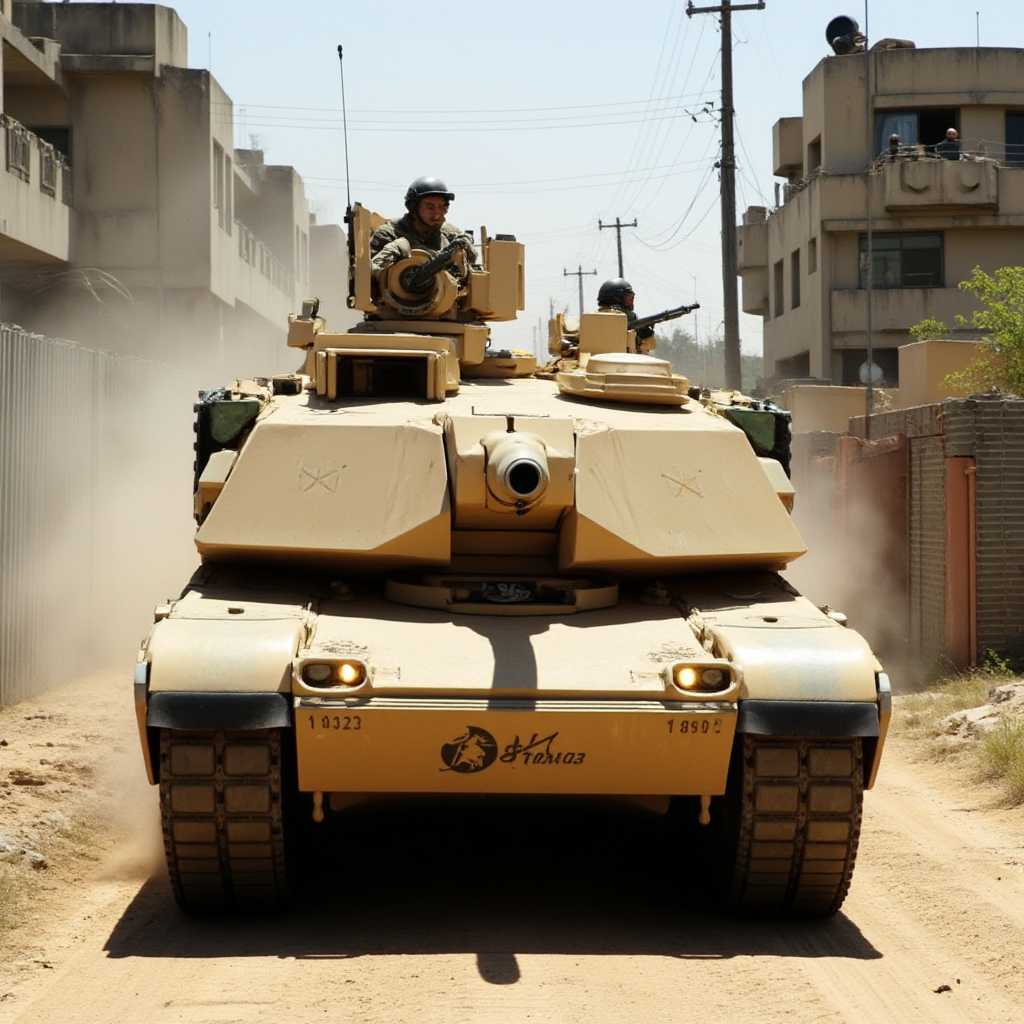}
        \caption{Generated by FLUX}
        \label{fig:abrams_flux}
    \end{subfigure}
    \hfill
    \hfill
    \begin{subfigure}[t]{0.3\textwidth}
        \centering
        \includegraphics[width=\linewidth]{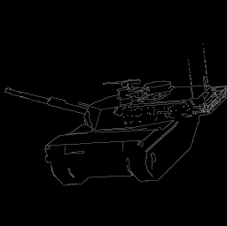}
        \caption{3D model–based edge map}
        \label{fig:abrams_cannyfg}
    \end{subfigure}
     \begin{subfigure}[t]{0.3\textwidth}
        \centering
        \includegraphics[width=\linewidth]{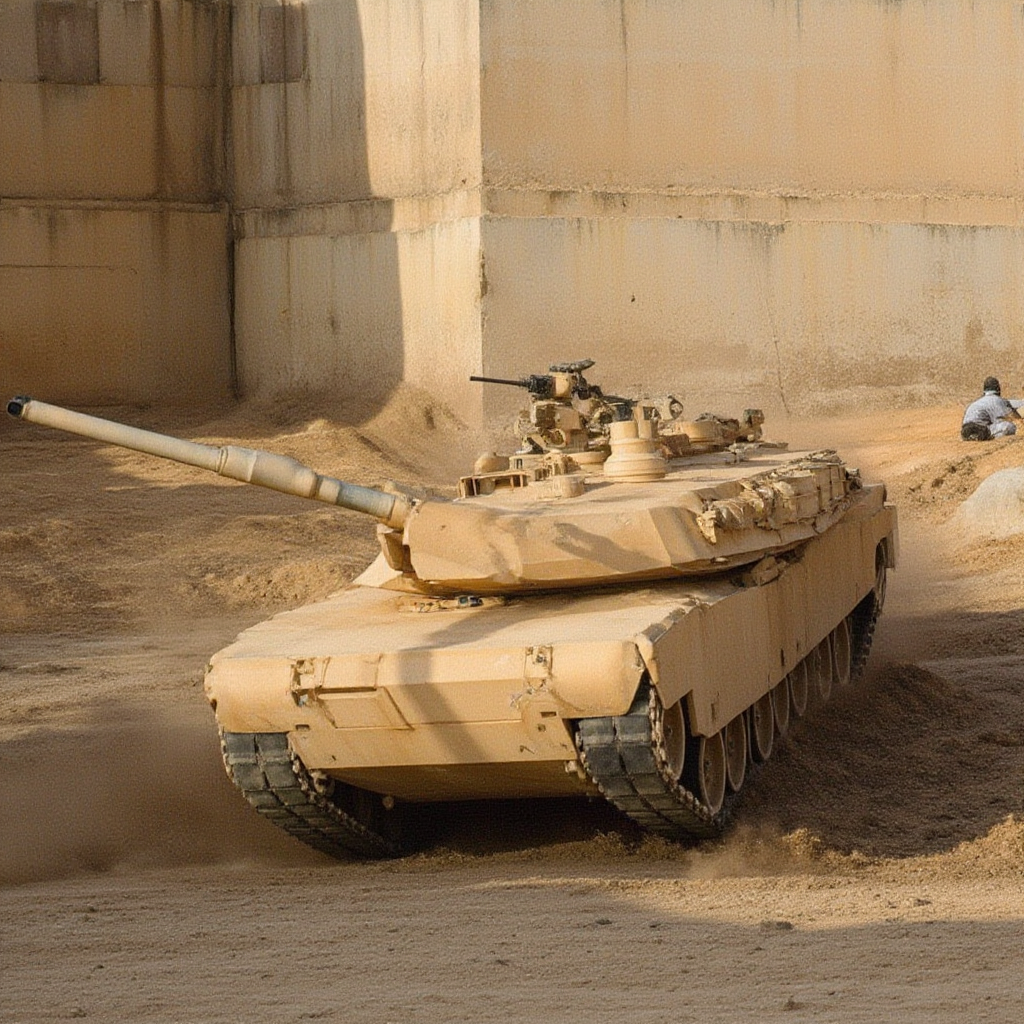}
        \caption{Generated by FLUX-CN}
        \label{fig:abrams_flux_cn}
    \end{subfigure}
    \caption{Top row: Boxer. Bottom row: M1A2 Abrams. From left to right: FLUX‑generated image, the corresponding 3D model–derived edge map used for ControlNet guidance, and the resulting FLUX‑CN image. ControlNet enables substantial additional variation in viewpoint and configuration while retaining realistic appearance.}
    \label{fig:examples_flux_vs_fluxcn}
\end{figure}

To address this, ControlNet-based structural guidance was introduced, using Canny edge maps derived from 3D vehicle models. Figure \ref{fig:c_flux_cn} shows the corresponding ControlNet-guided result for the Boxer example, and Figure \ref{fig:examples_flux_vs_fluxcn} provides additional comparisons for both the Boxer and M1A2 Abrams. The ControlNet‑guided samples demonstrate adherence to the intended spatial configuration, including viewpoint, orientation, and visible structure of the vehicle, while preserving the photorealistic appearance and environmental richness provided by FLUX. 

Table \ref{tab:flux_vs_fluxcn} reports detection performance when RF-DETR is fine-tuned on synthetic datasets generated either with \textit{FLUX} or with structurally guided \textit{FLUX-CN}. For the 8-sample setting, incorporating ControlNet guidance consistently improves performance across all configurations. When used as the sole synthetic data source, \textit{FLUX-CN} outperforms FLUX by $+4.1$ mAP$_{50}$. Similar gains are observed when synthetic data is combined with real images ($+5.1$) and when further combined with 3D model–based simulations ($+3.6$). These results indicate that structural guidance is particularly beneficial in very low-data regimes. For the 24-sample setting, ControlNet guidance does not lead to a clear performance improvement. Overall, the results show that ControlNet-based structural guidance improves the usefulness of generative data, when limited real data is available.

\begin{table}[tb!]
\centering
\caption{Comparison of RF-DETR performance when trained on FLUX vs.\ FLUX-CN synthetic data, with 8 or 24 real samples. 
Reported values are mAP$_{50}$ on the real test set (mean [$\pm$ std] over three runs). 
$\Delta$ indicates the difference in performance obtained by replacing FLUX with FLUX‑CN.}
\vspace{3mm}
\label{tab:flux_vs_fluxcn}
\begin{tabular}{llll}
\toprule
Training datasets RF-DETR  & FLUX       & FLUX-CN    & $\Delta$ (FLUX $\rightarrow$ FLUX-CN) \\  \midrule  \addlinespace[1mm] \textit{FLUX fine-tuned on 8 real samples} \\ \cmidrule{1-1}
synthetic only & 48.8 [2.0] & 52.9 [1.2] & $+4.1$                                \\
+ 8 real                & 70.9 [1.9] & 76.0 [2.7] & $+5.1$                                \\
+ 8 real + 3d-model-sim & 79.1 [0.6] & 82.7 [1.1] & $+3.6$ \\ \hline \addlinespace[1mm]      \textit{FLUX fine-tuned on 24 real samples} \\ \cmidrule{1-1}
synthetic only  & 57.7 [2.0] & 55.8 [2.2] & $-1.9$                                \\
+ 24 real                & 85.9 [1.1] & 88.2 [1.9] & $+2.3$                                \\
+ 24 real + 3d-model-sim & 88.4 [0.3] & 89.3 [1.5] & $+0.9$                                \\ \hline
\end{tabular}
\end{table}

\section{Discussion}
\label{sec:discussion}
% Summary 
% In this study, we investigated the effectiveness of generative AI-based synthetic data for improving military object detection in a 15-class setting. We generated synthetic datasets using the diffusion model FLUX, which was fine-tuned with LoRA on small sets of real images (8 and 24 samples) to incorporate domain-specific knowledge. To address the limited structural control of text-to-image diffusion models, we additionally generated structurally guided data using ControlNet, resulting in two types of synthetic datasets: FLUX and FLUX-CN. These datasets were used, both individually and in combination with real data and 3D-model-based simulations, to fine-tune an RF-DETR object detector. Performance was evaluated on a real-world test set, allowing us to assess (i) the extent to which generative synthetic data can improve detection performance, and (ii) whether explicit structural guidance during image synthesis provides additional benefit.

% Answer RQ1: GenAI adds value 
In this study, we demonstrate that synthetic data generated using diffusion models can effectively enhance military object detection performance. FLUX-based datasets consistently improve performance when combined with real training data. Compared to 3D model–based simulation, FLUX yields smaller performance gains. However, a key advantage of FLUX-based data generation is that it does not require 3D models, making it a promising alternative or complement to 3D model–based simulation pipelines. Combining all three data sources (real, 3D model–based, and generative AI) results in the highest performance, suggesting that these sources provide complementary information. While 3D-model simulations appear to provide stronger gains overall, likely due to improved geometric and viewpoint coverage, FLUX-based data contributes additional appearance variability that yields incremental improvements when combined with simulation.

When 24 real samples are available, the gain from \textit{FLUX} augmentation is small (approximately 3\%), and the gap between real+FLUX and real+FLUX+3D-simulation narrows (Figure \ref{fig:mAP_barchart}). This suggests that performance may be approaching saturation. However, it remains unclear whether this reflects an inherent upper bound on achievable performance (e.g. human-level performance has been reached), or whether the current synthetic data simply no longer provides additional benefit, leaving room for further improvement through more informative or higher-quality data.

% Answer RQ2: FLUX ControlNet works even better, especially for low number of simples (8) 
Incorporating structural guidance through ControlNet further improves the effectiveness of generative data, particularly in the low data domain. Visual inspection of the FLUX-CN-generated data indicates that combining generative models with explicit structural guidance enables more reliable control over geometric properties while maintaining high visual fidelity. For 8 real samples, FLUX-CN consistently outperforms FLUX when added to the fine-tuning dataset of RF-DETR, indicating that explicit control over object geometry and viewpoint enhances the usefulness of synthetic data when limited real examples are available. For 24 samples, the improvements are smaller and less consistent. 

While ControlNet improves structural fidelity, it currently relies on 3D model–derived edge maps, reintroducing the dependency on 3D assets. In addition, we observe a performance increase when FLUX-CN data is added on top of 3D model–based simulation data. This suggests that combining ControlNet‑guided generative data with traditional simulation is an effective and relatively low‑effort way to further boost performance, even when a 3D-simulation pipeline is already available.

\subsection{Limitations}
% Future work: applicability ControlNet to more challenging detection scenarios
While ControlNet improves control over geometry and viewpoint, this does not consistently translate into improved detection performance in the current setting when used in the 24-sample regime. A likely explanation is that the default generation biases of FLUX, such as frontal and close-range views, align relatively well with the real-image test distribution, whereas the increased geometric diversity introduced by FLUX-CN may lead to a larger domain gap.

% Future work: captioning, what works well? 
An open question is how to optimally distribute information between text prompts and structural conditioning. In the current setup, explicit geometric information is removed from the captions when using ControlNet, as this is provided through edge-based guidance. It has yet to be determined which attributes are best specified via text and which should be enforced through conditioning. 

% Discussion point: training regime
In the combined experiments, mixed datasets are obtained by repeating samples to get approximate ratios (50/50 for two sources and 33/33/33 for three sources), without explicitly enforcing these proportions per batch. While this provides a simple and effective setup, more systematic strategies for data mixing and sampling may further improve performance and should be explored in future work.

\subsection{Future work}
% Discussion point/future work: how do we get the edge maps? / another way of conditioning that does not depend on 3d models?
Maintaining control over object positioning and viewpoint without relying on explicit 3D models remains a challenge. Prior work has shown that diffusion models can be conditioned on a variety of structural signals, including edge maps, depth, segmentation, and pose \cite{zhang2023adding, lisanti2024conditioning}, as well as more coarse spatial constraints such as object layouts or bounding boxes \cite{bhat2024loosecontrol}. These directions suggest that it may be possible to obtain geometric control from alternative sources, for example derived from real images or learned representations, thereby reducing reliance on explicit 3D assets. Exploring such forms of guidance is an interesting direction for future work.

% Discussion point: blending in fore- and background -> related to decrease in recall 
In the current ControlNet setup, only edge maps of the foreground object are used (see Figures \ref{fig:b_canny_fg}, \ref{fig:abrams_cannyfg}, \ref{fig:boxer2_cannyfg}). This can lead to a mismatch in sharpness and detail between the object and the background, resulting in imperfect blending. In the present setting, this does not affect performance, as objects are typically large and easy to detect. However, in more challenging, most likely more military relevant scenarios with smaller or more distant objects, such artifacts may reduce robustness. Future work could explore alternative ways of incorporating background structure, for example by deriving those edges from real images.

% Discussion point: metrics and evaluation 
The difference between $\text{mAP}_{50}$ and $\text{mAP}_{50:95}$ ranges from 2.4 to 6.6 points across all settings (Table \ref{tab:mAP_flux}), which is smaller than typically reported for object detection benchmarks such as RF-DETR results on COCO \cite{gallagher2025rfdetr}. This indicates that localisation performance is already strong, and that performance gains from additional data sources primarily contribute to improved classification rather than improving bounding box accuracy or recall. This observation is intuitive for our use case, as the dataset predominantly contains close‑up views of vehicles, making object localisation relatively easy. At the same time, the task involves fine‑grained classification of visually similar vehicle classes, where classification is inherently more challenging than for coarse-grained benchmarks such as COCO. These findings highlight that mAP may obscure underlying error sources. More fine-grained analyses, such as error decomposition methods \cite{bolya2020tide}, could provide additional insight into whether improvements stem from classification, localisation, or missed detections. Furthermore, future work could focus on use cases where localisation is more challenging, for example scenarios with smaller objects, increased clutter, or larger camera–object distances.

\subsubsection{Conclusion}
Overall, the results demonstrate that generative AI-based synthetic data can effectively support military object detection in low-data settings. FLUX-based data consistently improves performance when combined with real images, and adding structural guidance via ControlNet provides additional benefits when limited real samples are available. These findings highlight the potential of object-specific diffusion models as a component in future military computer vision pipelines, complementing existing simulation approaches while reducing reliance on large curated real‑world datasets.

%\section*{ACKNOWLEDGMENTS} 

% References
\bibliography{report} % bibliography data in report.bib
\bibliographystyle{spiebib} % makes bibtex use spiebib.bst
\section{Appendix}
\label{appendix:prompts}
The system prompt that was provided to Gemma-3-12b-it, which was used to caption the real image dataset described in Section \ref{methods:datasets:real}, is reported in Listing \ref{caption_system_prompt}. The corresponding user prompt is reported in Listing \ref{caption_user_prompt}. The field $\{vehicle\_name\}$ is replaced by the vehicle class name. The system prompt that was provided to GPT-4, which was used for generating new captions as input for FLUX, is reported in Listing \ref{system_prompt}. The corresponding user prompt is reported in Listing \ref{user_prompt}. The $\{examples\_list\}$ consists of the captions generated using Gemma-3b-12it for the set of real training images.

\begin{lstlisting}[language=Python, caption=The system prompt used for real image caption generation with Gemma-3-12b-it.,label=caption_system_prompt]
You are a military vehicle image analysis specialist generating precise 
captions for AI image generation training.
Create detailed, natural language descriptions that would enable accurate 
recreation of military vehicle scenes.

CAPTION STRUCTURE:
Generate exactly two comprehensive sentences using active, descriptive 
language:

SENTENCE 1 - VEHICLE ANALYSIS: Detail the {vehicle_name} including all 
visible equipment, weaponry, cameras, camouflage 
pattern, operational condition, position&angle of visible canon, position 
of present turret, precise viewing perspective 
(front three-quarter, rear profile, side, front, rear three-quater, front 
above, side below, etc.), spatial positioning 
(close-up, medium distance, roadside, centered), and image quality 
characteristics (sharp focus, grainy texture, 
well-exposed, motion blur, etc.).

SENTENCE 2 - ENVIRONMENTAL CONTEXT: Describe how the vehicle interacts with 
its environment, including terrain surface, 
lighting conditions casting specific shadows, atmospheric effects (dust 
clouds, weather), background elements providing 
scale, and overall scene composition using active descriptive language.

TECHNICAL REQUIREMENTS:
- Use natural language as if explaining to a photographer
- Employ active verbs: "emerges," "cuts through," "reflects," "casts," 
"kicks up"
- Include specific camera/photo characteristics: "captured with telephoto 
compression," "shot at ground level," 
  "documentary-style framing"
- Describe interactions: "sunlight catches armor plating," "dust swirls 
around tracks," "shadows stretch across terrain"
- Specify technical details: camouflage patterns, equipment types, 
lighting direction, surface materials
- Note image quality: sharpness, contrast, exposure, grain, clarity
- Use precise military terminology naturally within descriptive sentences

OUTPUT: Two detailed, dynamic sentences only. No additional text.
\end{lstlisting}

\begin{lstlisting}[language=Python, caption=The user prompt used for real 
image caption generation with Gemma-3-12b-it.,label=caption_user_prompt]
You are a helpful assistant for generating creative prompts.
Always call the function 'return_prompts' and return ONLY a valid list of 
strings in the 'prompts' field.
\end{lstlisting}

\begin{lstlisting}[language=Python, caption=The system prompt used for synthetic caption generation with GPT-4.,label=system_prompt]
Create a comprehensive caption for this {vehicle_name} photograph using
natural, descriptive language that captures 
both the vehicle's technical details and its environmental context. 
Focus on active descriptions that would help an 
AI model accurately recreate this specific military scene, including 
image quality, spatial positioning, and atmospheric 
interactions.
\end{lstlisting}

\begin{lstlisting}[language=Python, caption=The user prompt used for synthetic caption generation with GPT-4.,label=user_prompt]
You are generating diverse, realistic military vehicle image captions for 
AI training. The vehicle type should remain consistent while varying all 
other elements to create authentic operational scenarios.

EXAMPLE TRAINING CAPTIONS:
{examples_list}

GENERATION REQUIREMENTS:
Create {batch} new captions that maintain the core vehicle identity while 
systematically varying (try to evenly spread all variations):

VEHICLE VARIATIONS (keep vehicle type consistent):
- Equipment loadouts: different mounted systems, communication arrays, armor 
configurations
- Operational condition: pristine, battle-worn, maintenance-ready, field-damaged, 
weathered
- Camouflage states: fresh paint, faded/worn, partially applied, mixed patterns, 
mud-covered

PERSPECTIVE & POSITIONING DIVERSITY:
- Viewing angles: (front, front quarter, rear three-quarter, side profile, rear) 
& (elevated view, ground-level, aerial perspective)
- Distance scales: extreme close-up, medium tactical distance, distant 
reconnaissance view
- Frame positioning: centered, off-center, edge of frame, partially obscured, 
dominant foreground

ENVIRONMENTAL SCENARIOS:
- Terrain variety: urban streets, desert sand, forest roads, muddy fields, rocky 
ground, snow-covered areas, combinations.
- Operational contexts: convoy movement, static guard duty, maintenance areas, 
combat zones, training exercises
- Weather conditions: clear conditions, rain/snow, dust storms, fog/haze, dramatic 
lighting

IMAGE QUALITY REALISM (mix of professional and field photography):
- High quality: crisp professional documentation, telephoto clarity, perfect 
exposure
- Standard field: handheld camera work, natural grain, adequate lighting, 
realistic focus
- Poor quality: surveillance footage, low resolution, overexposed/underexposed, 
motion blur, compression artifacts
- Mission documentation: night vision green, thermal imaging effects, 
helmet camera perspective, drone footage

ATMOSPHERIC INTERACTIONS:
Use active, descriptive language showing how elements interact:
- Light effects: "sunlight catches," "shadows stretch," "dust swirls," 
"reflections gleam"
- Environmental dynamics: "emerges through," "kicks up," "cuts across," 
"positioned against"
- Weather impacts: "rain streams down," "snow accumulates on," 
"heat shimmer distorts"

OUTPUT STYLE:
- Two comprehensive sentences per caption using natural, active language
- First sentence: vehicle description with technical details and positioning
- Second sentence: environmental context with atmospheric interactions
- Vary sentence structure and descriptive approaches
- Balance technical accuracy with readable, natural descriptions

Generate {batch} diverse captions now:

\end{lstlisting}

\end{document}